\documentclass[english]{scrartcl}
\usepackage{lmodern}

\usepackage[T1]{fontenc}
\usepackage[latin9]{inputenc}
\usepackage{geometry}
\usepackage{color}
\usepackage{babel}
\usepackage{array}
\usepackage{float}
\usepackage{multirow}
\usepackage{amsmath}
\usepackage{amsthm}
\usepackage{amssymb}
\usepackage{graphicx}
\usepackage[authoryear]{natbib}
\usepackage[unicode=true,pdfusetitle,
 bookmarks=true,bookmarksnumbered=false,bookmarksopen=false,
 breaklinks=false,pdfborder={0 0 0},pdfborderstyle={},backref=false,colorlinks=true]
 {hyperref}
\hypersetup{
 citecolor=blue}

\makeatletter

\providecommand{\tabularnewline}{\\}
\floatstyle{ruled}
\newfloat{algorithm}{tbp}{loa}
\providecommand{\algorithmname}{Algorithm}
\floatname{algorithm}{\protect\algorithmname}

\usepackage[font={small,it}]{caption}

\makeatother

\begin{document}

\title{Causal bootstrapping}

\author{Max A. Little$^{\star,\dagger}$, Reham Badawy$^{\star}$}
\maketitle
\begin{center}
\emph{$^{\star}$School of Computer Science, University of Birmingham,
UK}
\par\end{center}

\begin{center}
\emph{$^{\dagger}$MIT, Cambridge, MA, USA}
\par\end{center}

\begin{center}
\emph{First author contact: maxl@mit.edu}\footnote{\emph{This work partially funded by NIH grant UR-Udall Center, award
number P50 NS108676.}}
\par\end{center}
\begin{abstract}
To draw scientifically meaningful conclusions and build reliable engineering
models of quantitative phenomena, statistical models must take cause
and effect into consideration (either implicitly or explicitly). This
is particularly challenging when the relevant measurements are not
obtained from controlled experimental (interventional) settings, so
that cause and effect can be obscured by spurious, indirect influences.
Modern predictive techniques from machine learning are capable of
capturing high-dimensional, complex, nonlinear relationships between
variables while relying on few parametric or probabilistic modelling
assumptions. However, since these techniques are associational, applied
to observational data they are prone to picking up spurious influences
from non-experimental (observational) data, making their predictions
unreliable. Techniques from causal inference, such as probabilistic
causal diagrams and do-calculus, provide powerful (nonparametric)
tools for drawing causal inferences from such observational data.
However, these techniques are often incompatible with modern, nonparametric
machine learning algorithms since they typically require explicit
probabilistic models.

Here, we develop causal bootstrapping, a set of techniques for augmenting
classical nonparametric bootstrap resampling with information about
the causal relationship between variables. This makes it possible
to resample observational data such that, if it is possible to identify
an interventional relationship from that data, new data representing
that relationship can be simulated from the original observational
data. In this way, we can use modern machine learning algorithms unaltered
to make statistically powerful, yet causally-robust, predictions.
We develop several causal bootstrapping algorithms for drawing interventional
inferences from observational data, for classification and regression
problems, and demonstrate, using synthetic and real-world examples,
the value of this approach.
\end{abstract}

\section{Introduction}

One of the main aims of the quantitative sciences is to produce models
of observed phenomena in the world so that testable predictions can
be made from these models. In the ideal case, it is possible to perform
controlled experiments and measure the resulting change in variables
of interest. Data obtained from controlled experiments can be used
to produce models of the relationship between \emph{effects }and their
\emph{causes}, for example, modelling the relationship between treating
a medical issue by taking a drug (cause) and health status of that
issue (effect), while controlling for age which can be a cause of
both whether the drug is taken and health status \citep{pearl_causality:_2009}.
There are numerous situations where performing experiments is either
physically impossible, unethical or just infeasible from a practical
point of view. For example, we cannot modify the weather or prevent
the population of a country from using water to determine the causal
relationship between daily sunshine hours and domestic water usage,
independent of atmospheric humidity which influences both cloud cover
and the need for people to hydrate themselves.

Broadly speaking, there are two kinds of data about the world: \emph{experimental
}data obtained from controlled experiments, and \emph{observational
}data. In many sciences such as medicine and agriculture, the \emph{randomized
controlled trial }is the archetypal experiment used to test the effectiveness
of a treatment \citep{matthews_introduction_2006}. Well-developed
statistical analysis of this data allows us to quantify the strength
of the causal effect of the treatment on the outcome of interest.
Whereas, in epidemiology and economics, data is almost entirely observational
since experiments are usually impractical -- a key question therefore
for these mainly observational sciences is whether it is possible
to nonetheless infer causes and their effects from the available data.

A variety of ``tricks'' for statistical analysis of such data have
been developed to address this problem, among them so-called \emph{adjustment
methods }and \emph{instrumental variables}. However, these tricks
only work in special circumstances. Whether and how these tricks can
be generalized to address a wider range of observational data raises
important questions which have, over the last few decades, coalesced
into the new discipline of \emph{causal inference}. Arguably, the
most systematic, complete and integrated work in this discipline uses
the tools of (\emph{probabilistic})\emph{ causal diagrams}, \emph{do-calculus
}and other conceptual and analytical devices \citep{pearl_causality:_2009}.

Meanwhile, new forms of experimental and observational data have become
usable due to advances in digital measurement, storage and processing
hardware. For example, it is now possible to capture, store and process
millions of digitized X-ray images recording the presentation of various
medical conditions in the medical clinic. Digital measurements of
patterns of human transport or online behaviour using devices such
as smartphones, have also become available. This data is nearly all
observational. As with data from ``classical'' observational epidemiological
or ecological settings, it is generally infeasible (in terms of cost
and/or logistics) to run controlled experiments to determine causal
relationships in this setting. However, it differs from classical
settings in that (1) the data is of enormous scale (on the order of
billions of observations across thousands or millions of variables
is not unusual), and (2) there are unknown, complex (nonlinear) rather
than simple linear, relationships between these variables. Analyzing
this kind of data to make reliable causal inferences is a challenge
for traditional statistics but the discipline of \emph{statistical
machine learning} has emerged to take on some of these challenges.
Machine learning predictors, such as kernel regression, random forests,
support vector machines and deep learning, have been developed to
learn high-dimensional, nonlinear statistical relationships between
variables \citep{little_machine_2019}.

While these machine learning predictors can have extremely high accuracy,
the major drawback is their complete ``blindness'' to causal structure.
This is because they find \emph{associational }relationships, not
causal ones. For example, empirical evidence points to these predictors
easily exploiting spurious associations in observational data \citep{zech_variable_2018,chyzhyk_controlling_2018,kaufman_leakage_2012,neto_permutation_2019,neto_using_2018}.
Thus, these machine learning methods cannot learn causal (interventional)
relationships from observational data (at least not without special
adaptations). It would therefore be valuable to somehow co-opt the
predictive power of these machine learning algorithms, meanwhile,
ensuring they can make interventional predictions from the available
observational data, without having to make special adaptations to
these algorithms.

Here, we introduce a simple method to achieve this. We augment the
classical \emph{bootstrap }resampling method \citep{efron_introduction_1994}
with information from the causal diagram generating the observational
data. This leads to a simple \emph{weighted }bootstrap which can be
used to generate new data faithful to an interventional distribution
of interest. Any standard, complex nonlinear machine learning predictor
can then be applied to the new data to construct \emph{interventional
predictors}, rather than associational predictors. This method is
applicable to most interventional distributions which can be derived
from observational causal models using the rules of do-calculus, according
to the general identification algorithm of \citet{shpitser_complete_2008}.

We develop several bootstrap algorithms for common causal inference
scenarios including general back-door and front-door deconfounding,
tailored to supervised classification or regression machine learning
methods. We demonstrate the effectiveness of this technique for synthetic
data and real-world, practical causal inference problems.

\section{Methods}

First we introduce some notation. Labels such as $X$, $Y$, $U$,
$W$ and $Z$ refer to random variables and their realizations, $x,y,u,w$
and $z$. These have sample spaces $\Omega_{X},\Omega_{Y}$, and so
on. Multidimensional variables are bold, e.g. $\boldsymbol{X}$ and
$\boldsymbol{x}$, and $x_{n}$ is the $n$-th observation of the
variable $x$. For the set of variables $\mathcal{S}=\left\{ x,y,u,w\right\} $,
the shorthand $p\left(z,\mathcal{S}\right)$ refers to the joint probability
density (PDF) or probability mass function (PMF) $p\left(z,x,y,u,w\right)$,
and $\mathcal{S}_{n}$ refers to tuple of observations $\left(x_{n},y_{n},u_{n},w_{n}\right)$.
The notation $\int p\left(z,\mathcal{S}\right)d\mathcal{S}$ refers
to the marginalization of the variables $\mathcal{S}$ from $p$.
A causal graph, which is a \emph{directed acyclic graphical }(DAG)
model with vertices $\mathcal{V}$ being a set of random variables,
indicates the conditional independence relationships between the variables.
The edges in the graph capture dependencies, e.g. $U\leftarrow Y\to\boldsymbol{X}$
encodes that both $U$ and $\boldsymbol{X}$ depend upon $Y$, but
$Y$ does not depend upon either. The set-valued function $\mathcal{P}:\mathcal{V}\to2^{\mathcal{V}}$
gives the set of \emph{parent }variables for $U$, e.g. if $\mathcal{P}\left(u\right)=\left\{ v,w,z\right\} $,
then $p\left(u|\mathcal{P}\left(u\right)\right)=p\left(u|v,w,z\right)$,
representing these relationships.

Our approach uses nonparametric estimates of interventional distributions,
and manipulates them analytically to produce a simple expression capturing
the causal relationship of interest, from which interventional samples
can be simulated from the observational data. We make use of nonparametric
\emph{kernel density estimates }(KDE) obtained from joint and marginal
\emph{reproducing kernel Hilbert space }(RKHS) functions $K\left[\cdot\right]$,
to illustrate: 
\begin{eqnarray}
\hat{p}\left(x,y\right) & = & \frac{1}{N}\sum_{n\in\mathcal{N}}K\left[x-x_{n}\right]K\left[y-y_{n}\right]\label{eq:joint-KDE}\\
\hat{p}\left(x\right) & = & \frac{1}{N}\sum_{n\in\mathcal{N}}K\left[x-x_{n}\right]\label{eq:univariate-KDE}
\end{eqnarray}
where $\mathcal{N}=\left\{ 1,2,\ldots,N\right\} $ is the set of indices
of the sample data for the random variables $X,Y$. Note that the
kernels for $X$ and $Y$ can generally be distinct. Simplifying these
KDEs to relies on two basic mathematical devices. The first is the
\emph{reproducing property} \citep{berlinet_reproducing_2011}:
\begin{equation}
\left\langle p,K\left[x,\cdot\right]\right\rangle =\int p\left(x^{\prime}\right)K\left[x-x^{\prime}\right]dx^{\prime}=p\left(x\right)
\end{equation}

Here, $p\left(x\right)$ is some PDF or PMF, and the integral computes
a marginal. Using this property, we analytically solve the marginal
integral by replacing occurrences of the variable $x$, with evaluations
of the distribution $p\left(x_{n}\right)$ at the realization $x_{n}$.
The second property is the linearity of integration allowing us to
swap integrals with summations.

Through such analytical manipulations, the ``heart'' of our approach
involves simple \emph{weighted }interventional KDEs:
\begin{equation}
p\left(\boldsymbol{x}|do\left(y\right)\right)\approx\sum_{n\in\mathcal{N}}K\left[\boldsymbol{x}-\boldsymbol{x}_{n}\right]w_{n}\label{eq:weighted-KDE}
\end{equation}
where $0<w_{n}<\infty$ is some real-valued weighting vector. Sampling
from this equation is straightforward; we draw some $i\in\mathcal{N}$
with probability proportional to $w_{i}$, then draw a value $\boldsymbol{x}$
from the kernel function $K$ centered on $\boldsymbol{x}_{i}$. Furthermore,
if we replace the kernel $K$ in this equation with the Dirac delta
function (for continuous $\boldsymbol{X}$), then the value $\boldsymbol{x}=\boldsymbol{x}_{i}$
without the need to sample from the kernel. This is the basis of the
\emph{bootstrap} \citep{efron_introduction_1994}, which motivates
the description of sampling from models like (\ref{eq:weighted-KDE})
as \emph{causal bootstrapping}.

\begin{figure}
\centering{}\includegraphics[scale=0.6]{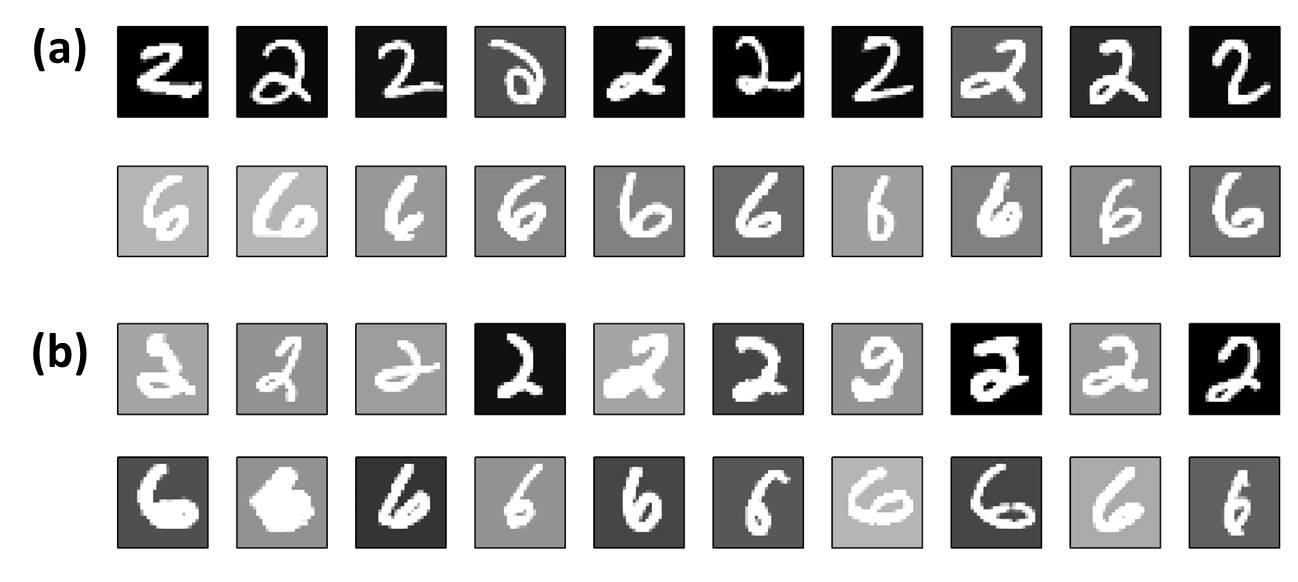}\caption{Using causal bootstrapping to construct an interventional machine
learning predictor for digit recognition when the observational data
is confounded. The ``brightness-MNIST'' problem has variables where
$U$ is the image brightness, $\boldsymbol{X}$ are the pixels and
$Y$ is the target digit label, (`2' vs. `6'). (a) In the observational
data, brightness $U$ is a confounder since it is a common cause of
the target and the image data, $Y\leftarrow U\to\boldsymbol{X}$.
Therefore, it is highly correlated with the digit label $Y$, in which
case, any standard supervised classifier trained on this data has
a high risk of simply using the brightness to predict the digit. (b)
When the digit label and brightness are independent (which would be
the case in an experiment where the digit label is controlled) then
there is no advantage to using the brightness information to make
a digit label prediction. However, in practice, the observational
training data may be confounded as in (a). Back-door (Algorithm \ref{alg:back-door-bootstrap})
or (when there is a mediator) front-door causal bootstrapping (Algorithm
\ref{alg:front-door-bootstrap}) can be used to simulate the controlled
experiment (b), on which a classifier which correctly solves the problem
of digit label recognition, can be trained. Random forest classification
accuracy trained on confounded data (a) reaches 96\% out-of-sample,
but tested on non-confounded data (b) collapses to near chance, 58\%.
By contrast, training on causal bootstrapped data achieves greater
than 90\% tested out-of-sample and similar accuracy on (b). See Table
\ref{tab:classifiers-backdoor-frontdoor} for details.\label{fig:background-MNIST}}
\end{figure}
Here we illustrate briefly an application of these ideas to supervised
machine learning. Consider that we want to learn the nonlinear relationship
$Y,\boldsymbol{X}$. Here, $Y$ is a univariate \emph{prediction target}
variable and $\boldsymbol{X}$ is a high-dimensional \emph{feature
}variable. However, there is a variable $U$ which introduces the
\emph{confounding path }$Y\leftarrow U\to\boldsymbol{X}$ (Figure
\ref{fig:causal-graphs}a). This means we cannot, from the training
data pairs $\mathcal{D}=\left(\boldsymbol{x}_{n},y_{n}\right)$, $n\in\mathcal{N}$,
learn the \emph{causal }relationship\emph{ }which would have been
obtained by measuring the training data under an experiment whereby
we control $Y$ independent of $U$. This is because the data was
generated from $p\left(\boldsymbol{x}|y\right)$ which is not the
same as $p\left(\boldsymbol{x}|do\left(y\right)\right)$ due to the
confounding path. If the resulting predictor, trained on $\left(\boldsymbol{x}_{n},y_{n}\right)$
were to be used in a situation in which there was no confounding,
we cannot expect it to make reliable predictions. However, if we have
also measured $u_{n}$, we can resample a new, \emph{deconfounded
}set of training data $\mathcal{D}^{\star}$ by causal bootstrapping
from (\ref{eq:weighted-KDE}) using the weights:

\begin{equation}
w_{n}=\frac{K\left[y_{n}-y\right]}{N\,\hat{p}\left(y|u_{n}\right)}
\end{equation}
where $K=\mathbf{1}$, the discrete Kronecker delta, in the classification
case where where the sample space $Y\in\Omega_{Y}$ is discrete, or
$K$ is some suitable kernel in the (univariate) regression case where
$\Omega_{Y}=\mathbb{R}$ is continuous. The conditional $\hat{p}\left(y|u\right)$
can obtained using any suitable density estimator (we suggest using
KDEs for their simplicity). In the classification setting where $Y,U$
are both discrete, this method is \emph{parameter-free, }and thus
the only source of additional error in this bootstrap procedure over
and above those sources in the original data, is due to bootstrap
resampling variability alone.
\begin{algorithm}[H]
\textbf{Input}: $N$ samples $\mathcal{D}=\left(\boldsymbol{x}_{n},y_{n},\mathcal{S}_{n}\right)$,
$n\in\mathcal{N}=\left\{ 1,2,\ldots,N\right\} $ from a graphical
model, and samples from the back-door admissible variable set $\mathcal{S}$.
The variables are: arbitrary multidimensional feature data (vector)
$\boldsymbol{X}$, prediction target $Y$ and arbitrary adjustment
set $\mathcal{S}$, with sample spaces $\Omega_{\boldsymbol{X}},\Omega_{Y}$
and $\Omega_{\mathcal{S}}$.

\textbf{Output}: $N$ deconfounded samples $\mathcal{D}^{\star}=\left(\boldsymbol{x}_{m},y_{m}\right)$,
$m\in\mathcal{N}$ approximating samples from $p\left(\boldsymbol{x}|do\left(y\right)\right)$.
\begin{enumerate}
\item Find empirical KDEs $\hat{p}\left(y,\mathcal{S}\right)$ and $\hat{p}\left(\mathcal{S}\right)$
from $\mathcal{D}$ to compute $\hat{p}\left(y|\mathcal{S}\right)$.
\item For each $n\in\mathcal{N}$:
\item $\quad$Produce new sample $\mathcal{D}^{\star}=\left(\boldsymbol{x}_{i},y_{n}\right)$,
where index $i$ is selected from $\mathcal{N}$ with weights:
\[
w_{i}=\frac{K\left[y_{i}-y_{n}\right]}{N\,\hat{p}\left(y_{n}|\mathcal{S}_{i}\right)}
\]
where $K\left[\cdot\right]=\mathbf{1}\left[\cdot\right]$ for classification,
and a suitable kernel for regression.
\end{enumerate}
\caption{Back-door causal bootstrapping for supervised classification and regression.\label{alg:back-door-bootstrap}}
\end{algorithm}

In this way, we can co-opt any high-performance, predictive machine
learning algorithm to learn the desired causal relationship from observational
training data without the need to modify the machine learning algorithm,
nor perform a separate, potentially logistically difficult, controlled
experiment. See Figure (\ref{fig:background-MNIST}) for an example
of high-dimensional machine learning prediction from confounded digital
image data, deconfounded using causal bootstrapping.

\subsection{Bootstrap weights for interventional distributions}

\begin{figure}
\centering{}\includegraphics[scale=0.6]{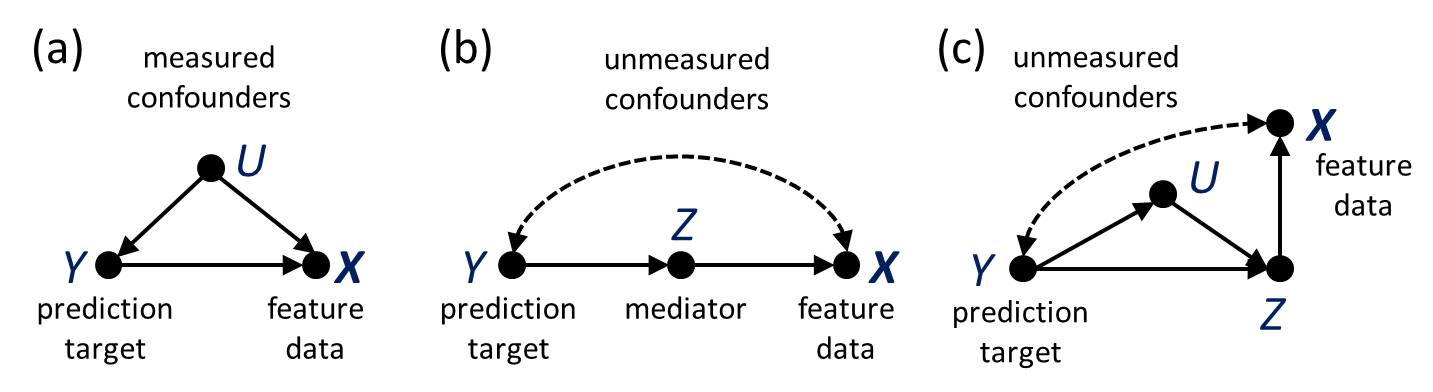}\caption{Causal graphs for the supervised prediction relationship $Y,\boldsymbol{X}$
in the presence of various sources of confounding and mediation. (a)
A simple special case of back-door confounding through the measured
variable(s) $U$, for which Algorithm \ref{alg:back-door-bootstrap}
can be used to bootstrap deconfounded data consistent with intervening
on $Y$. (b) Front-door confounding, where the confounding is unmeasured
but there exists a mediator $Z$. Here Algorithm \ref{alg:front-door-bootstrap}
can be used to deconfound observational data from this graph. (c)
A special graph used to illustrate the generality of causal bootstrapping.\label{fig:causal-graphs}}
\end{figure}

We now develop the theoretical justification for causal bootstrapping
such as Algorithm \ref{alg:back-door-bootstrap}. In causal inference
problems, we often have an interventional distribution in the form:
\begin{equation}
p\left(\boldsymbol{x}|do\left(y\right)\right)=\int p\left(\boldsymbol{x}|\mathcal{P}\left(\boldsymbol{x}\right)\right)\prod_{v\in\mathcal{E}}p\left(v|\mathcal{P}\left(v\right)\right)d\mathcal{E}\label{eq:intervention-distrib}
\end{equation}
where $\boldsymbol{X}$ is (primary) effect variable, $Y$ is the
intervention variable (prediction target), and $\mathcal{E}=\mathcal{P}\left(\boldsymbol{x}\right)\backslash y$
are secondary effect variables which are marginalized out. This can
be obtained by various methods such as truncated factorization, back-
or front-door deconfounding, or the general identification algorithm
of \citet{shpitser_complete_2008}. The causal bootstrapping weights
in (\ref{eq:weighted-KDE}) to simulate from this distribution, given
observational data, are given by:

\begin{eqnarray}
w_{n} & = & \frac{1}{N}\times\begin{cases}
K\left[y_{n}-y\right]\bar{w}_{n} & \textrm{if }y\in\mathcal{P}\left(\boldsymbol{x}\right)\\
\bar{w}_{n} & \textrm{otherwise}
\end{cases}\label{eq:intervention-kernel}\\
\bar{w}_{n} & = & \left.\frac{\prod_{v\in\mathcal{E}}\hat{p}\left(v|\mathcal{P}\left(v\right)\right)}{\hat{p}\left(\mathcal{P}\left(\boldsymbol{x}\right)\right)}\right|_{\forall u\in\mathcal{E}:u=u_{n}}\label{eq:intervention-weights}
\end{eqnarray}

In the second line, each occurrence of the $u\in\mathcal{E}$ is replaced
with the realization $u_{n}$ from the observational data (note that
$y\notin\mathcal{E}$ by construction of interventional distributions).
For example, if $p\left(v|\mathcal{P}\left(v\right)\right)=p\left(v|x,z,y\right)$,
then $p\left(v|\mathcal{P}\left(v\right)\right)=p\left(v_{n}|x_{n},z_{n},y\right)$
after replacement. The notation $\hat{p}$ refers to a (generally
nonparametric) estimate of the true PDF/PMF, $p$. A proof of this
is given in \nameref{sec:Appendix-A}.

As an example of the above, given a causal graphical model, the \emph{identification
}(ID) algorithm of \citet{shpitser_complete_2008} can be used to
determine whether any desired interventional distribution can be obtained
from this model. If so, it produces the expression for the interventional
distribution. Using this expression and the interventional formula
(\ref{eq:intervention-distrib})-(\ref{eq:intervention-weights}),
we can simplify this to a causal bootstrap (with the only restriction
that $\mathcal{E}=\mathcal{P}\left(\boldsymbol{x}\right)\backslash y$).
As an example, \citet{tikka_identifying_2017} derive the following
formula for the causal effect of $Y$ on $\boldsymbol{X}$ from the
causal diagram of Figure \ref{fig:causal-graphs}c:

\begin{equation}
p\left(\boldsymbol{x}|do\left(y\right)\right)=\int p\left(w\right)\int p\left(z|w,y\right)\int p\left(\boldsymbol{x}|w,y^{\prime},z\right)p\left(y^{\prime}|w\right)dy^{\prime}dz\,dw
\end{equation}

Following the above and applying RKHS estimators and then simplifying,
leads to the following weighted KDE:

\begin{equation}
f\left(\boldsymbol{x}|do\left(y\right)\right)\approx\frac{1}{N}\sum_{n\in\mathcal{N}}K\left[\boldsymbol{x}_{n}-\boldsymbol{x}\right]\frac{\hat{p}\left(z_{n}|w_{n},y\right)}{\hat{p}\left(z_{n}|w_{n},y_{n}\right)}\label{eq:tikka-KDE}
\end{equation}
For a proof of this, see \nameref{sec:Appendix-A}.

\subsection{Choice of effect kernel}

The choice of RKHS kernel function for the effect $\boldsymbol{X}$
 depends heavily on the sample spaces of the variables involved.
For instance, if $\Omega_{X}$ is discrete, it makes sense to use
the Kronecker delta $\mathbf{1}\left[x\right]=1$ for $x=0$, zero
otherwise. For continuous sample spaces, the choice depends to a large
extent on computational or smoothness considerations; many kernels
have \emph{bandwidth }parameters which control the regularity of the
KDE for the variable \citep{silverman_density_1986}.

We will also make use the \emph{Dirac delta }function as a ``kernel''
for $\boldsymbol{X}$; although this is not in an RKHS, it does satisfy
the reproducing property, e.g. $\int p\left(x^{\prime}\right)\delta\left[x-x^{\prime}\right]dx^{\prime}=p\left(x\right)$
which is in fact all we need. Replacing $K=\delta$ in (\ref{eq:weighted-KDE})
leads to an even more tractable bootstrap model. To sample from this
model, on having chosen $i$ as described above, it suffices to simply
emit the sample $\boldsymbol{x}_{i}$ (since all the mass of the estimator
is concentrated on the point set $\boldsymbol{x}_{n}$, $n\in\mathcal{N}$).
This is the basis of bootstrap\emph{ }resampling \citep{efron_introduction_1994}.

Next, we develop the application of the theory above to several examples
from causal inference, and derive associated causal bootstrapping
algorithms.

\subsection{Back-door causal bootstrap}

A common situation encountered in practice is that of \emph{confounding}
where unwanted causal paths exist between the (prediction target)
variable $Y$ and the observed feature data $\boldsymbol{X}$, interfering
with the direct causal path of interest relating $Y$ to $\boldsymbol{X}$.
It is possible to estimate the interventional distribution $p\left(\boldsymbol{x}|do\left(y\right)\right)$
if an \emph{admissible set }of variables\emph{ }$\mathcal{S}$, can
be found \citep{pearl_causality:_2009}. This set must satisfy the
\emph{back-door criterion}: (i) no variable in $\mathcal{S}$ is a
descendent of $Y$, and (ii) the variables $\mathcal{S}$ \emph{block}
all causal paths with an arrow pointing to $Y$. Applying the rules
of do-calculus shows that the interventional distribution can be obtained
using: 
\begin{equation}
p\left(\boldsymbol{x}|do\left(y\right)\right)=\int p\left(\boldsymbol{x}|y,\mathcal{S}\right)p\left(\mathcal{S}\right)d\mathcal{S}
\end{equation}

Now, using (\ref{eq:intervention-kernel})-(\ref{eq:intervention-weights}),
we obtain the following \emph{back-door adjusted }KDE:

\begin{equation}
p\left(\boldsymbol{x}|do\left(y\right)\right)\approx\frac{1}{N}\sum_{n\in\mathcal{N}}K\left[\boldsymbol{x}-\boldsymbol{x}_{n}\right]\frac{K\left[y_{n}-y\right]}{\hat{p}\left(y|\mathcal{S}_{n}\right)}\label{eq:back-door-KDE}
\end{equation}
A detailed derivation is given in \nameref{sec:Appendix-A}. An interesting
special admissible set are the \emph{direct parents }of $Y$, that
is, the variables upon which $Y$ depends immediately in the DAG \citep{pearl_causality:_2009}.
By definition they cannot be descendents of $Y$ (satisfying criterion
(i)) and since all back-door paths must go through the incoming edges
to $Y$, these can only originate in the direct parents of $Y$ (satisfying
condition (ii)). Another example is shown in Figure \ref{fig:causal-graphs}a
with a single back-door path blocked by the variable $U$. Selecting
$K\left[\boldsymbol{x}-\boldsymbol{x}_{n}\right]=\delta\left[\boldsymbol{x}-\boldsymbol{x}_{n}\right]$,
leads to the simple \emph{back-door causal bootstrap }algorithm which
is suitable for supervised classification and regression applications,
Algorithm \ref{alg:back-door-bootstrap}.

Note that in this algorithm, if $Y$ is discrete, rather than simulating
a single intervention for each sample $y_{n}$ in $\mathcal{D}$,
we can usually simplify the computations by grouping together all
simulated data that share the same value of $y$. If, for each $y\in\Omega_{Y}$
we simulate $\left\lfloor N\,\hat{p}\left(y\right)\right\rfloor $
samples with the same value of $y$, we ensure that both the observed
$\hat{p}\left(y\right)$ and the number of observations in $\mathcal{D}$
are retained in $\mathcal{D}^{\star}$. However, this choice of the
distribution of resampled $Y$ is not a requirement for the back-door
causal bootstrap to be valid. If $Y$ is continuous, the values $y_{n}$
are all distinct and cannot be grouped to simplify the computations.
Instead, the observed distribution $\hat{p}\left(y\right)$ is reproduced
exactly by simulating a single intervention for each sample $y_{n}$.
However, as with the discrete case, we do not have to reproduce the
observed, marginal distribution of $Y$ when generating deconfounded
data; indeed we can simulate any interventional dataset we wish. For
example, when $N$ is very large, it may be more practical to produce
a smaller, deconfounded dataset across a uniformly sampled, representative
range of values of $Y$ instead.

\subsection{Front-door causal bootstrap}

A somewhat more complex situation that arises in some observational
settings is that of so-called \emph{front-door }confounding. Here,
multiple back-door paths composed of unobserved variables exist between
the prediction target $Y$ and the observed feature data $\boldsymbol{X}$,
interfering with the causal relationship $Y,\boldsymbol{X}$ (Figure
\ref{fig:causal-graphs}b). Since we cannot observe these back-door
variables, we cannot block using back-door deconfounding. However,
there is also a variable $Z$, known as a \emph{mediator}, such that
$Y\to Z\to\boldsymbol{X}$. In this situation, we can use do-calculus
to derive an expression for $p\left(\boldsymbol{x}|do\left(y\right)\right)$
such that we do not need to explicitly block any of the back-door
paths.

The front-door criterion is in three parts: (i) the mediator must
intercept all paths between $Y$ and $\boldsymbol{X}$, (ii) all back-door
paths from $Z$ to $\boldsymbol{X}$ must be blocked by $Y$, and
(iii) there should be no other paths between $Z$ and $\boldsymbol{X}$.
If these conditions hold, we get the following interventional distribution
\citep{pearl_introduction_2010}:

\begin{equation}
p\left(\boldsymbol{x}|do\left(y\right)\right)=\int\left(\int p\left(\boldsymbol{x}|y^{\prime},z\right)p\left(y^{\prime}\right)dy^{\prime}\right)p\left(z|y\right)dz
\end{equation}

As with back-door deconfounding, using the interventional formula
(\ref{eq:intervention-distrib})-(\ref{eq:intervention-weights})
leads to the following \emph{front-door adjusted }KDE (see \nameref{sec:Appendix-A}
for derivation):

\begin{equation}
f\left(\boldsymbol{x}|do\left(y\right)\right)\approx\frac{1}{N}\sum_{n\in\mathcal{N}}K\left[\boldsymbol{x}_{n}-\boldsymbol{x}\right]\frac{\hat{p}\left(z_{n}|y\right)}{\hat{p}\left(z_{n}|y_{n}\right)}\label{eq:front-door-KDE}
\end{equation}

This gives us the \emph{front-door causal bootstrap}, Algorithm (\ref{alg:front-door-bootstrap}),
which we describe for supervised machine learning applications. As
with the back-door algorithm, the computations may be simplified by
looping over each $y\in\Omega_{Y}$ in the discrete interventional
variable case.\textbf{}

\begin{algorithm}[H]
\textbf{Input}: $N$ marginal samples $\mathcal{D}=\left(\boldsymbol{x}_{n},y_{n},z_{n}\right)$,
$n\in\mathcal{N}=\left\{ 1,2,\ldots,N\right\} $ from the front-door
confounded causal graphical model. The variables are: arbitrary feature
data (vector) $\boldsymbol{X}$, prediction target $Y$ and mediator
$Z$, with sample spaces $\Omega_{\boldsymbol{X}},\Omega_{Y}$ and
$\Omega_{Z}$.

\textbf{Output}: $N$ deconfounded samples $\mathcal{D}^{\star}=\left(\boldsymbol{x}_{m},y_{m}\right)$,
$m\in\mathcal{N}$ approximating samples from $p\left(\boldsymbol{x}|do\left(y\right)\right)$.
\begin{enumerate}
\item Find empirical EDF $\hat{p}\left(z,y\right)$ from $\mathcal{D}$
to compute $\hat{p}\left(z|y\right)$.
\item For each $n\in\mathcal{N}$:
\item $\quad$Produce new sample $\mathcal{D}^{\star}=\left(\boldsymbol{x}_{i},y_{n}\right)$,
where index $i$ is selected from $\mathcal{N}$ with weights:
\[
w_{i}=\frac{\hat{p}\left(z_{i}|y_{n}\right)}{N\,\hat{p}\left(z_{i}|y_{i}\right)}
\]
\end{enumerate}
\caption{Front-door causal bootstrapping for supervised classification and
regression problems.\label{alg:front-door-bootstrap}}
\end{algorithm}

\subsection{Truncated factorization causal bootstrap}

More generally, for a causal graphical model $\mathcal{G}$ with vertices
$\mathcal{V}=\left\{ \boldsymbol{x},\mathcal{E},y\right\} $ and $\mathcal{E}=\mathcal{P}\left(\boldsymbol{x}\right)\backslash y$,
the joint distribution is given by:

\begin{eqnarray}
p\left(\boldsymbol{x},\mathcal{E},y\right) & = & p\left(\boldsymbol{x}|\mathcal{P}\left(\boldsymbol{x}\right)\right)p\left(y|\mathcal{P}\left(y\right)\right)\prod_{v\in\mathcal{E}}p\left(v|\mathcal{P}\left(v\right)\right)
\end{eqnarray}

If all the variables are observed, we can use the \emph{truncated
factorization }formula to compute the interventional distribution
\citep{pearl_introduction_2010}:

\begin{eqnarray}
p\left(\boldsymbol{x},\mathcal{E}|do\left(y\right)\right) & = & p\left(\boldsymbol{x}|\mathcal{P}\left(\boldsymbol{x}\right)\right)\prod_{v\in\mathcal{E}}p\left(v|\mathcal{P}\left(v\right)\right)
\end{eqnarray}
Marginalizing out other effect variables $\mathcal{E}$ isolates the
causal effect of $Y$ on $\boldsymbol{X}$:
\begin{eqnarray}
p\left(\boldsymbol{x}|do\left(y\right)\right) & = & \int p\left(\boldsymbol{x}|\mathcal{P}\left(\boldsymbol{x}\right)\right)\prod_{v\in\mathcal{E}}p\left(v|\mathcal{P}\left(v\right)\right)d\mathcal{E}
\end{eqnarray}

This is in the form of interventional distribution (\ref{eq:intervention-distrib}),
and it follows that the causal bootstrap weights for $p\left(\boldsymbol{x}|do\left(y\right)\right)$
are those given by plugging in the KDEs $\hat{p}\left(\mathcal{P}\left(\boldsymbol{x}\right)\right)$
and $\hat{p}\left(v|\mathcal{P}\left(v\right)\right)$ for all $v\in\mathcal{E}$
and into (\ref{eq:intervention-kernel}). This leads to Algorithm
\ref{alg:truncated-bootstrap} suitable for supervised learning applications.
As above, the computations may be simplified by looping over each
for $y\in\Omega_{Y}$ in the discrete interventional variable case.

\begin{algorithm}[H]
\textbf{Input}: $N$ samples $\mathcal{D}=\left(\boldsymbol{x}_{n},\mathcal{E}_{n},y_{n}\right)$,
$n\in\mathcal{N}=\left\{ 1,2,\ldots,N\right\} $ from the joint distribution
over the graphical causal model $\mathcal{G}$ with variables $\mathcal{V}=\left\{ \boldsymbol{x},\mathcal{E},y\right\} $,
where$\boldsymbol{X}$ is an arbitrary multidimensional feature data
(vector), $Y$ is the prediction target, and arbitrary additional
variables $\mathcal{E}$, with sample spaces $\Omega_{\boldsymbol{X}},\Omega_{Y}$
and $\Omega_{\mathcal{E}}$.

\textbf{Output}: $N$ samples $\mathcal{D}^{\star}=\left(\boldsymbol{x}_{m},y_{m}\right)$,
$m\in\mathcal{N}$ approximating the interventional distribution $p\left(\boldsymbol{x}|do\left(y\right)\right)$.
\begin{enumerate}
\item Using $\mathcal{D}$, find empirical KDEs $\hat{p}\left(v|\mathcal{P}\left(v\right)\right)$
, for all $v\in\mathcal{E}$, and the joint KDE $\hat{p}\left(\mathcal{P}\left(\boldsymbol{x}\right)\right)$.
\item For each $n\in\mathcal{N}$:
\item $\quad$Produce new sample $\mathcal{D}^{\star}=\left(\boldsymbol{x}_{i},y_{n}\right)$,
where index $i$ is selected from $\mathcal{N}$ with weights $w_{i}$
given by:
\begin{eqnarray*}
w_{i} & = & \frac{1}{N}\times\begin{cases}
K\left[y_{i}-y_{n}\right]\bar{w}_{i} & \textrm{if }y\in\mathcal{P}\left(\boldsymbol{x}\right)\\
\bar{w}_{i} & \textrm{otherwise}
\end{cases}\\
\bar{w}_{i} & = & \left.\frac{\prod_{v\in\mathcal{E}}\hat{p}\left(v|\mathcal{P}\left(v\right)\right)}{\hat{p}\left(\mathcal{P}\left(\boldsymbol{x}\right)\right)}\right|_{\forall u\in\mathcal{E}:u=u_{i}}
\end{eqnarray*}
where $K\left[\cdot\right]=\mathbf{1}\left[\cdot\right]$ for discrete
prediction target $Y$, or another suitable kernel otherwise.
\end{enumerate}
\caption{Truncated factorization causal bootstrapping for supervised classification
and regression.\label{alg:truncated-bootstrap}}
\end{algorithm}

\section{Experiments and results}

Next, we run several numerical experiments to demonstrate the application
of the above algorithms in practice. For full details of these experiments,
see \nameref{sec:Appendix-B}.

\subsection{Synthetic Gaussian mixtures}

\begin{figure}
\centering{}\includegraphics[viewport=10cm 10cm 150cm 80cm,clip,scale=0.1]{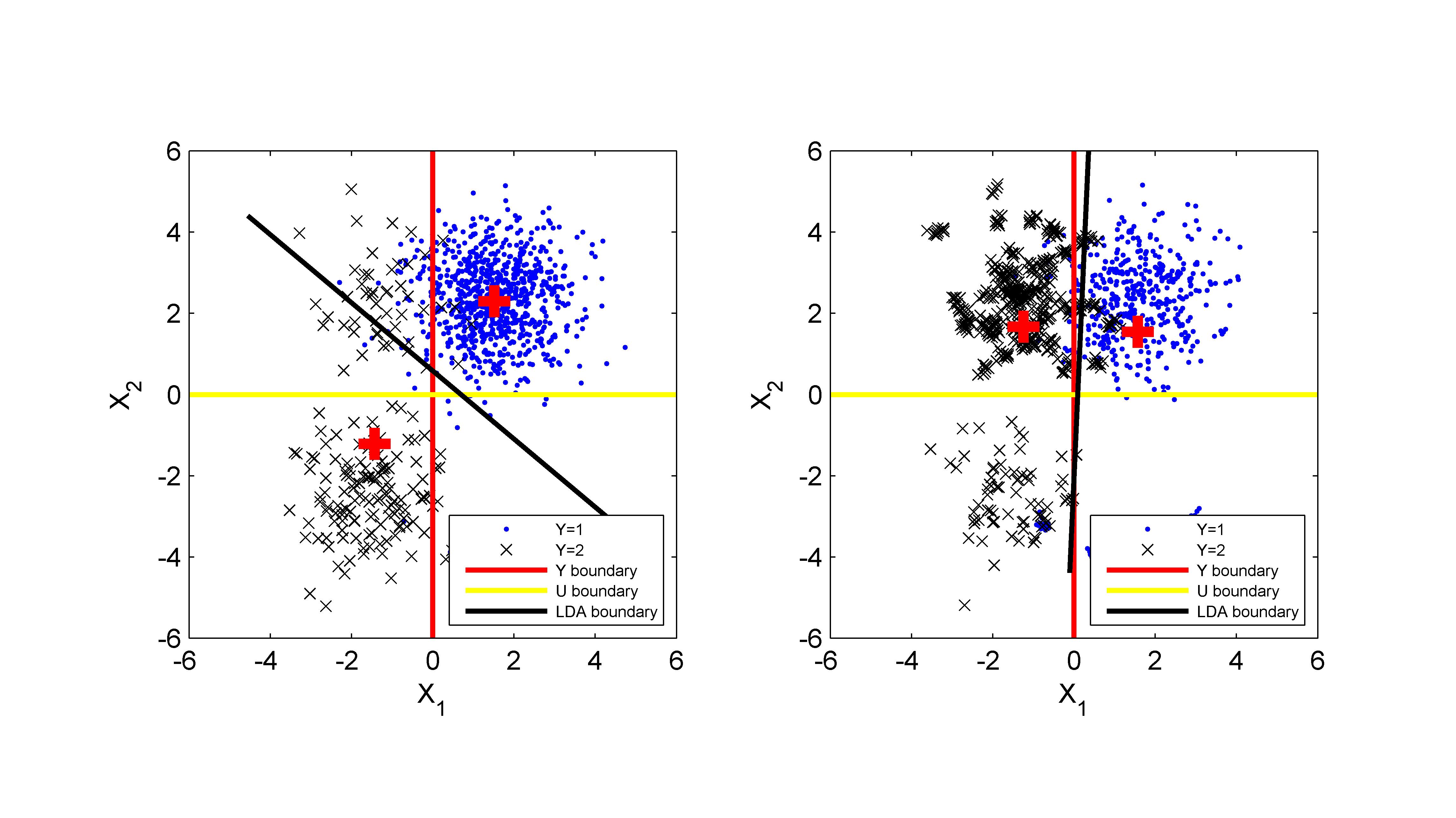}\caption{Synthetic example illustrating causal bootstrapping for classification.
\emph{Left}: bivariate Gaussian feature data $\boldsymbol{X}$ depends
upon both the discrete confounder $U$, and the discrete classification
target (red versus black points) $Y$. The target $Y$ also depends
upon the confounder. In this situation, the ideal boundary to correctly
classify on the basis of the target is vertical, whereas, the confounded
boundary is horizontal. The black line is the boundary which would
be determined using linear discriminant analysis (LDA). \emph{Right}:
data after applying back-door or front-door causal bootstrapping.
The resampling causes the LDA boundary to almost entirely coincide
with the correct classification target $Y$ boundary.\label{fig:back-door-simple}}
\end{figure}

In this section, we demonstrate a simple, confounded model involving
bivariate Gaussian features and discrete targets, confounder and mediator
(Figure \ref{fig:back-door-simple}). We apply back-door (Algorithm
\ref{alg:back-door-bootstrap}) and (when there is a mediator) front-door
causal bootstrapping in order deconfound the data. Trained on the
deconfounded data, a simple linear discriminant (LDA) classifier achieves
typical accuracies of 85-95\% on both confounded and deconfounded
data (Table \ref{tab:classifiers-backdoor-frontdoor}). This easily
outperforms LDA trained on the original, confounded data and tested
on the deconfounded data where reaches at best 73\% (back-door) and
collapses to chance accuracy in the front-door case. Figure \ref{fig:back-door-simple}
gives us an intuitive explanation for how causal bootstrapping works
in these cases: it changes the density of samples in feature space
such that the confounded boundary is dominated by the desired target
boundary instead.

\begin{table}
\begin{centering}
\begin{tabular}{|>{\raggedright}m{2cm}|>{\raggedright}m{3.5cm}|>{\centering}p{3cm}|>{\centering}p{2cm}|>{\centering}p{2cm}|}
\cline{3-5} 
\multicolumn{1}{>{\raggedright}m{2cm}}{} &  & \emph{Training data}\\
Sample 1 & \emph{Test data}\\
Sample 2 (confounded) & \emph{Test data}\\
Sample 3 (non-confounded)\tabularnewline
\hline 
\multirow{6}{2cm}{\emph{Back-door}} & \multirow{2}{3.5cm}{\emph{Synthetic Gaussian mixture (LDA)}} & Confounded & $97\pm1$ & $73\pm3$\tabularnewline
\cline{3-5} 
 &  & Deconfounded & $95\pm2$ & $\mathbf{91\pm1}$\tabularnewline
\cline{2-5} 
 & \multirow{2}{3.5cm}{\emph{Background- MNIST (RF)}} & Confounded & $96\pm1$ & $58\pm5$\tabularnewline
\cline{3-5} 
 &  & Deconfounded & $97\pm2$ & $\mathbf{93\pm2}$\tabularnewline
\cline{2-5} 
 & \multirow{2}{3.5cm}{\emph{Parkinson's voice (RF)}} & Confounded & $90\pm2$ & $57\pm3$\tabularnewline
\cline{3-5} 
 &  & Deconfounded & $73\pm7$ & \textbf{$\mathbf{69\pm3}$}\tabularnewline
\hline 
\multirow{4}{2cm}{\emph{Front-door}} & \multirow{2}{3.5cm}{\emph{Synthetic Gaussian mixture (LDA)}} & Confounded & $98\pm1$ & $50\pm1$\tabularnewline
\cline{3-5} 
 &  & Deconfounded & $85\pm2$ & $\mathbf{84\pm1}$\tabularnewline
\cline{2-5} 
 & \multirow{2}{3.5cm}{\emph{Background- MNIST (RF)}} & Confounded & $94\pm2$ & $52\pm3$\tabularnewline
\cline{3-5} 
 &  & Deconfounded & $94\pm1$ & $\mathbf{86\pm}3$\tabularnewline
\hline 
\end{tabular}
\par\end{centering}
\caption{Accuracy of classifiers (linear discriminant analysis, LDA, and random
forests, RF) applied to both confounded sample data, and sample data
deconfounded using back-door and front-door causal bootstrapping (Algorithms
\ref{alg:back-door-bootstrap}-\ref{alg:front-door-bootstrap}). Classes
are balanced so that chance predictions have 50\% accuracy. Data samples
1 \& 2 are confounded, whereas data sample 3 is non-confounded. When
making predictions for non-confounded data, classifiers trained on
deconfounded data clearly outperform classifiers trained on the original,
confounded data (bold figures). By contrast, training on the original,
confounded data produces classifiers which perform very poorly on
non-confounded data, often no better than chance. However, training
and testing on confounded data shows high test classification accuracy,
giving a misleading indication of performance in practice where the
data is not confounded.\label{tab:classifiers-backdoor-frontdoor}}
\end{table}

\subsection{Semi-synthetic ``background-MNIST''}

In these examples, we simulate measured (back-door) or unmeasured/unknown
(front-door) confounding in an image recognition problem. Here, a
set of MNIST digits are modified such that the brightness is altered
depending upon the digit. The task is to recognize the digit. More
specifically, the brightness is altered in a way which depends upon
the digit label. This makes the brightness a strong confounder, and
this clear signal is likely to be exploited by any predictor instead
of actually detecting the configuration of pixels in the image (Figure
\ref{fig:background-MNIST}). The setup is similar to the situation
in medical imaging which sometimes occurs, where the settings of the
imaging machine are inadvertently linked to the disease category which
is to be classified \citep{zech_variable_2018}. A random forest (RF)
classifier trained on the causal bootstrapped data performs well across
both confounded and non-confounded data, whereas the same classifier
trained on the original, confounded training data performs well on
the confounded test data, but on the non-confounded data it collapses
down to nearly chance performance (Table \ref{tab:classifiers-backdoor-frontdoor}).

\subsection{Real-world data: Parkinson's voice}

In this final example, we test back-door causal bootstrapping on real-world
experimental data captured from individuals with Parkinson's disease
(PD). The experimental goal is to detect, from digital voice recordings,
whether the individual has been diagnosed with PD or not. Nine voice
features are extracted from the recordings, which are then input to
a classifier. Three separate experiments have been conducted under
varying conditions (different labs in different countries, speakers
of different languages, slightly different age and sex grouping).
In order to improve statistical estimation, the aim is to merge the
data from the experiments together. However, the differences in experimental
conditions mean that the distribution of the features differ. At the
same time, there is an imbalance in the number of recordings collected
from each experiment. Combined, these cause confounding meaning that
a classifier may get a spurious advantage from detecting the experimental
setup rather than PD/healthy status. This is a significant problem
with voice-based disease characterization \citep{little_using_2017}.

Indeed, we find that an RF classifier trained on the confounded data,
whilst performing well on an independent, confounded test set, performs
significantly worse than an RF classifier trained on the same data
deconfounded using back-door causal bootstrapping (Table \ref{tab:classifiers-backdoor-frontdoor}).
However, the performance of the classifier on the confounded data
is much higher, strong evidence that the classifier relies heavily
on the confounding signal in order to make classification decisions.

\section{Related work}

Causal modelling analysis is well established across certain disciplines,
but the introduction of causal graphical models with probabilistic
foundations led to defining causal phenomena such as interventions,
confounding, collider and mediation variables in one unified framework
\citep{pearl_causality:_2009}. The awareness that (supervised) machine
learning algorithms are prone to spurious confounding, and what to
do about it, has a relatively long history, but it is only recently
that rigorous solutions are being proposed, see \citet{kaufman_leakage_2012}.
More recently, \citet{zech_variable_2018,voets_replication_2019}
examine confounding in the context of deep learning for high-dimensional
predictors. \citet{little_using_2017} discussed proposals for identification
of confounding involving cross-validation stratified on the confounding
variable. Similarly, to detect confounding, \citet{neto_using_2018}
suggested the use of stratified permutation testing and \citet{ferrari_measuring_2019}
use train/test sets stratified on both prediction target and confounder.
These studies do not detail a method for constructing deconfounded
predictors and are restricted to the problem of classification under
a single measured confounding variable (see Figure \ref{fig:causal-graphs}a).

The problem has only very recently started to be expressed in explicitly
causal terms. Use of the back-door adjustment formula with explicit
probabilistic prediction models has been proposed in \citet{landeiro_robust_2016}
and \citet{landeiro_controlling_2017}. These methods require an explicit
probabilistic model for the predictor and so do not apply to all supervised
machine learning algorithms. Perhaps closest to this work, \citet{chyzhyk_controlling_2018}
propose importance sampling to create deconfounded data from observational
data. However, this does not guarantee that the marginal distributions
of the confounder and target variables are retained which is problematic
because a predictor can be sensitive to these marginals. Furthermore,
it can only create test sets which are a subset of the training data.
It is restricted to the situation with a single measured confounder
as in Figure \ref{fig:causal-graphs}a.

Related methods for constructing interventional predictors are now
being explored in the machine learning community. For example, \citet{kallus_removing_2018}
investigate the special situation where a small, interventional sample
is available which overlaps with a larger observational study, where
the overlap contains sufficient information to construct an interventional
predictor. The approach is not fully nonparametric, relying on a parametric
correction of an predictor constructed from the observational data.
A broader approach invokes \emph{causal stability} \citep{pearl_causality:_2009},
that is, constructing predictors which only exploit information that
does not, or is not expected to, change across different observational
settings. For example, \emph{invariant causal prediction }methods
combine data from multiple interventional samples; using this they
select, or more generally find a representation of, the variables
that are parents of the prediction target \citep{peters_causal_2016,arjovsky_invariant_2019}
to construct a predictor which is invariant across differing experimental
or observational settings. These methods rely upon the existence of
multiple datasets collected under different settings. By contrast,
if information about which parts of the causal graph may change is
available, \citet{subbaswamy_preventing_2019} present an algorithm
for identifying a stable interventional distribution (if one can be
found).

Meanwhile, the causal inference community have recently begun to explore
the use of machine learning prediction methods to augment ``standard''
causal inference methods such as confounder stratification, propensity
score matching and inverse probability weighting. \citet{athey-2018}
developed a new model for the average cause-effect relationship stratified
by confounders, where the stratification borrows the efficient subdivision
strategy of random forests \citep{little_machine_2019}. Similarly,
predictive machine learning algorithms have also been used to reduce
bias and variance in estimating average treatment effects using deconfounding
methods \citep{schuler_2017}. Unlike causal bootstrapping, they only
apply to causal estimation in the simplest setting of Figure \ref{fig:causal-graphs}a.

As discussed earlier, causal bootstrapping is an \emph{entirely general
}method for causal inference, because it allows simulation of the
complete interventional distribution from observational data, if such
interventional distribution can be computed. For example, it is straightforward
to perform \emph{anti-causal estimation}, where causes are inferred
from effects, as is common in classification in machine learning (see
the examples above).

Also, some standard causal inference methods are special cases of
causal bootstrapping, as shown next. Taking the causal bootstrap empirical
PDF for binary treatments, $p\left(\boldsymbol{x}|do\left(y\right)\right)\approx$\linebreak{}
$\sum_{n\in\mathcal{N}}K\left[\boldsymbol{x}-\boldsymbol{x}_{n}\right]w_{n}$,
the average causal effect $\boldsymbol{X}$ given the intervention
$Y=y$ is:

\begin{equation}
\begin{aligned}E_{\boldsymbol{X}}\left[\boldsymbol{X}|do\left(y\right)\right]\approx & \sum_{n\in\mathcal{N}}w_{n}\int\boldsymbol{x}\,K\left[\boldsymbol{x}-\boldsymbol{x}_{n}\right]d\boldsymbol{x}\\
= & \sum_{n\in\mathcal{N}}w_{n}\boldsymbol{x}_{n}
\end{aligned}
\end{equation}

In the case of back-door deconfounding (Algorithm \ref{alg:back-door-bootstrap})
and binary interventions $Y\in\left\{ 0,1\right\} $, inserting the
associated weights in the above leads to:

\begin{align}
E_{\boldsymbol{X}}\left[\boldsymbol{X}|do\left(y\right)\right] & \approx\frac{1}{N}\sum_{n\in\mathcal{N}}\boldsymbol{x}_{n}\frac{1\left[y_{n}-y\right]}{\hat{p}\left(y|u_{n}\right)}
\end{align}
which is the classical inverse probability weighting (IPW) estimator
for the average treatment effect (ATE). Indeed, the generality of
causal bootstrapping can lead to useful new ATEs, for example, for
instance in the front-door confounding situation (Algorithm \ref{alg:front-door-bootstrap})
the expected response is:
\begin{align}
E_{\boldsymbol{X}}\left[\boldsymbol{X}|do\left(y\right)\right] & \approx\frac{1}{N}\sum_{n=1}^{N}\boldsymbol{x}_{n}\frac{\hat{p}\left(\boldsymbol{z}_{n}|y\right)}{\hat{p}\left(\boldsymbol{z}_{n}|y_{n}\right)}
\end{align}
This is a novel probability weighting estimator for deconfounding
ATEs under \emph{mechanisms }$\boldsymbol{Z}$ and \emph{unmeasured
}confounding covariates. Similar formulae for causal mediation settings
such as this have more recently been exhibited but involve much lengthier
and opaque derivations than given above \citep{fulcher_2019}. It
should be clear that similar logic as above applied to (\ref{eq:intervention-kernel})-(\ref{eq:intervention-weights})
leads to an entirely novel class of ATE estimators. All these estimators
could even be modified to create robust \emph{median} or \emph{quantile
}causal effect estimates suitable for situations where the average
is not a good summary statistic \citep{little_machine_2019}.

\section{Discussion and conclusions}

In this paper we have shown how, given a structural causal model and
observational data from that model, it is possible to derive simple
algorithms to draw bootstrap samples from that data which are consistent
with a desired interventional distribution. We have developed several
algorithms for supervised machine learning, which may be of general
usefulness, including those based on closing back-door confounding
paths, unmeasured and unknown confounding satisfying the front-door
criterion, and more generally, an algorithm for any given interventional
distribution. Through empirical experiments, we have demonstrated
that effective interventional predictors can be trained using this
technique that do not merely make associational predictions. At the
same time, these result highlight the striking negative impact of
failure to take confounding properly into account in supervised machine
learning.

There are certain limitations to this approach. In particular, we
need to know the causal graphical model for the observational data.
This can be difficult for situations where there is significant ambiguity.
However, in many situations we do not need to know all the causal
relationships, and large parts of the graph may be unknown. Consider
for instance front-door deconfounding: provided the criteria hold,
we do not need to know about, or measure, all the variables on the
back-door causal paths which are intercepted by the target $Y$. This
makes front-door causal bootstrapping fairly generic as a deconfounding
tool. It may also be possible to use structural discovery algorithms
and hence recover an estimate of the causal model for the observational
data \citep{chickering_optimal_2002}.

Another problem with this approach is where the distribution of a
variable needs to be estimated yet it is high-dimensional, for example
the distribution $\hat{p}\left(y|\mathcal{S}\right)$ in Algorithm
\ref{alg:back-door-bootstrap}. Although we have suggested using KDEs,
these estimates can be unreliable when there is a paucity of data
available. However, we note that the causal bootstrap algorithms proposed
here do not stipulate the form of these distribution estimates, and
parametric estimates may be more reliable, particularly if additional
information about these variables is available.

Bootstrapping necessarily involves producing repeated observations
of the effect variable $\boldsymbol{X}$. In some applications this
is problematic. For example, given an observational sample, we may
want to test the performance of a causal predictor by bootstrapping
multiple interventional samples from that original data. However,
these multiple samples will share individual observations with the
original data, this will introduce an optimistic bias into the performance
estimates, which is particularly problematic for modern complex predictors
such as deep learning that can memorize individual observations \citep{belkin_reconciling_2018}.
One solution to this is to use bootstrap bias correction methods \citep[Section 7.11]{hastie_elements_2009}.
An alternative is to use split-sample approaches for example, mimicking
cross-validation, whereby the original observational sample is bootstrapped
once and the bootstrap is split into multiple subsamples. The predictor
is trained and tested on non-overlapping subsamples of the bootstrap
such that no observations are shared between train and test.

Although we have generally emphasised the use of the Dirac delta function
for the effect variable $\boldsymbol{X}$ in the causal bootstrap
algorithms, a different direction is to use a kernel other than the
delta function. This naturally leads to the causal analogue of \emph{smoothed
bootstrapping} \citep{silverman_bootstrap:_1987}. One advantage of
such smoothed bootstraps is that all causal bootstraps consist of
genuinely unique observations (assuming $\boldsymbol{X}$ is continuous)
not shared with the original observational sample. This will mitigate
the above problem of observation memorization for out-of-sample performance
testing. However, it is often the case that $\boldsymbol{X}$ will
be high-dimensional so that the smoothed bootstraps may deviate quite
substantially from the underlying (and unknown) distribution of $\boldsymbol{X}$,
if the observational sample is not sufficiently large.

Finally, we note that the RKHS-based KDE for the joint distribution
of $\boldsymbol{X}$ and its parents, utilizing the reproducing property
to derive the bootstrap weights, is not the only RKHS-based estimator.
For example, we can also use semi-parametric RKHS \emph{finite }or
\emph{infinite mixture models }\citep{little_machine_2019} which
may be preferable in certain situations, for example when the data
is naturally clustered and the size of observational data is too small
to guarantee reliable bootstrap samples. We leave the derivation of
the corresponding causal bootstrap algorithms for future work.

\bibliographystyle{plainnat}
\phantomsection\addcontentsline{toc}{section}{\refname}\bibliography{Causal_Bootstrap}

\section*{Appendix A: Proofs\label{sec:Appendix-A}}

\addcontentsline{toc}{section}{Appendix A: Proofs}

\subsection*{Bootstrap weights for interventional distributions}

Here we prove that the causal bootstrapping weights in (\ref{eq:weighted-KDE})
from an interventional distribution where the set of unwanted (secondary)
effect variables $\mathcal{E}=\mathcal{P}\left(\boldsymbol{x}\right)\backslash y$,
are given by (\ref{eq:intervention-kernel})-(\ref{eq:intervention-weights}).
Starting with the interventional distribution and expanding out the
conditional $p\left(\boldsymbol{x}|\mathcal{P}\left(\boldsymbol{x}\right)\right)$:

\begin{equation}
p\left(\boldsymbol{x}|do\left(y\right)\right)=\int p\left(\boldsymbol{x},\mathcal{P}\left(\boldsymbol{x}\right)\right)\frac{\prod_{v\in\mathcal{E}}p\left(v|\mathcal{P}\left(v\right)\right)}{p\left(\mathcal{P}\left(\boldsymbol{x}\right)\right)}d\mathcal{E}\label{eq:intervention-expand}
\end{equation}
We now replace $p\left(\boldsymbol{x},\mathcal{P}\left(\boldsymbol{x}\right)\right)$
with the RKHS KDE:
\begin{equation}
p\left(\boldsymbol{x},\mathcal{P}\left(\boldsymbol{x}\right)\right)\approx\frac{1}{N}\sum_{n\in N}K\left[\boldsymbol{x}-\boldsymbol{x}_{n}\right]\prod_{u\in\mathcal{P}\left(\boldsymbol{x}\right)}K\left[u-u_{n}\right]\label{eq:x-parent-KDE}
\end{equation}

Inserting this into (\ref{eq:intervention-expand}), and since $\boldsymbol{x}\notin\mathcal{E}$,
we can factorize the multiple integral: 
\begin{eqnarray}
p\left(\boldsymbol{x}|do\left(y\right)\right) & \approx & \frac{1}{N}\sum_{n\in N}K\left[\boldsymbol{x}-\boldsymbol{x}_{n}\right]\int\prod_{u\in\mathcal{P}\left(\boldsymbol{x}\right)}K\left[u-u_{n}\right]\frac{\prod_{v\in\mathcal{E}}p\left(v|\mathcal{P}\left(v\right)\right)}{p\left(\mathcal{P}\left(\boldsymbol{x}\right)\right)}d\mathcal{E}
\end{eqnarray}
such that the weights (\ref{eq:intervention-kernel}) are given by:

\begin{equation}
w_{n}=\frac{1}{N}\int\prod_{u\in\mathcal{P}\left(\boldsymbol{x}\right)}K\left[u-u_{n}\right]\frac{\prod_{v\in\mathcal{E}}p\left(v|\mathcal{P}\left(v\right)\right)}{p\left(\mathcal{P}\left(\boldsymbol{x}\right)\right)}d\mathcal{E}
\end{equation}

Now, picking one $u^{\prime}\in\mathcal{E}$, the integral above can
be further factorized:

\begin{equation}
\int\prod_{u\in\mathcal{P}\left(\boldsymbol{x}\right)\backslash u^{\prime}}K\left[u-u_{n}\right]\int K\left[u^{\prime}-u_{n}^{\prime}\right]\frac{\prod_{v\in\mathcal{E}}f\left(v|\mathcal{P}\left(v\right)\right)}{f\left(\mathcal{P}\left(\boldsymbol{x}\right)\right)}du^{\prime}d\left(\mathcal{E}\backslash u^{\prime}\right)
\end{equation}
and evaluating the inner integral using the reproducing property we
get:

\begin{equation}
\int K\left[u^{\prime}-u_{n}^{\prime}\right]\frac{\prod_{v\in\mathcal{E}}p\left(v|\mathcal{P}\left(v\right)\right)}{p\left(\mathcal{P}\left(\boldsymbol{x}\right)\right)}du^{\prime}=\left.\frac{\prod_{v\in\mathcal{E}}p\left(v|\mathcal{P}\left(v\right)\right)}{p\left(\mathcal{P}\left(\boldsymbol{x}\right)\right)}\right|_{u^{\prime}=u_{n}^{\prime}}
\end{equation}
that is, every instance of the variable $u^{\prime}$ on the left
hand side is replaced by the realization $u_{n}^{\prime}$ on the
right. This same pattern of factorization and evaluation of integrals
by replacement is repeated for all $u^{\prime}\in\mathcal{E}$. Also,
every distribution $p$ is replaced by its estimate from the data
$\hat{p}$. This proves the form of the vector $\bar{w}_{n}$ in (\ref{eq:intervention-weights}).
Finally, since $Y$ is the intervention variable it is not integrated
out, so that if $y\in\mathcal{P}\left(\boldsymbol{x}\right)$ the
kernel $K\left[y-y_{n}\right]$ is retained in (\ref{eq:x-parent-KDE}),
thus it also features in computing the weights $w_{n}$, proving (\ref{eq:intervention-kernel}).

\subsection*{Bootstrap weights for causal algorithms}

Here we derive the back-door weighted KDE (\ref{eq:back-door-KDE})
using (\ref{eq:intervention-kernel})-(\ref{eq:intervention-weights}).
Here $\mathcal{E}=\left\{ \mathcal{S}\right\} $, the interventional
parents are $\mathcal{P}\left(\boldsymbol{x}\right)=\left\{ y,\mathcal{S}\right\} $
(so $y\in\mathcal{P}\left(\boldsymbol{x}\right)$), and $\mathcal{P}\left(\mathcal{S}\right)=\emptyset$,
giving $w_{n}=N^{-1}K\left[y-y_{n}\right]\bar{w}_{n}$ with:

\begin{eqnarray}
\bar{w}_{n} & = & \frac{\hat{p}\left(\mathcal{S}_{n}\right)}{\hat{p}\left(y,\mathcal{S}_{n}\right)}=\frac{1}{\hat{p}\left(y|\mathcal{S}_{n}\right)}
\end{eqnarray}
and substituting these $w_{n}$ into (\ref{eq:weighted-KDE}) gives
(\ref{eq:back-door-KDE}).

Similarly, for the front-door formula (\ref{eq:front-door-KDE}),
we have $\mathcal{E}=\left\{ y^{\prime},z\right\} $, interventional
distribution parents $\mathcal{P}\left(\boldsymbol{x}\right)=\left\{ y^{\prime},z\right\} $
(so $y\notin\mathcal{P}\left(\boldsymbol{x}\right)$), $\mathcal{P}\left(y^{\prime}\right)=\emptyset$,
$\mathcal{P}\left(z\right)=\left\{ y\right\} $, leading to:

\begin{eqnarray}
w_{n} & = & \frac{\hat{p}\left(y_{n}\right)\hat{p}\left(z_{n}|y\right)}{N\,\hat{p}\left(y_{n},z_{n}\right)}=\frac{\hat{p}\left(z_{n}|y\right)}{N\,\hat{p}\left(z_{n}|y_{n}\right)}
\end{eqnarray}
and substituting these $w_{n}$ into (\ref{eq:weighted-KDE}) gives
(\ref{eq:front-door-KDE}).

Finally, we prove the form of the weights (\ref{eq:tikka-KDE}) for
the interventional distribution given in \citet{tikka_identifying_2017}.
Here, $\mathcal{E}=\left\{ y^{\prime},z,w\right\} $, with interventional
parents $\mathcal{P}\left(\boldsymbol{x}\right)=\left\{ w,y^{\prime},z\right\} $
(implying that $y\notin\mathcal{P}\left(\boldsymbol{x}\right)$),
$\mathcal{P}\left(y^{\prime}\right)=\left\{ w\right\} $, $\mathcal{P}\left(z\right)=\left\{ w,y\right\} $
and $\mathcal{P}\left(w\right)=\emptyset$ giving:

\begin{eqnarray}
w_{n} & = & \frac{\hat{p}\left(y_{n}|w_{n}\right)\hat{p}\left(z_{n}|w_{n},y\right)\hat{p}\left(w_{n}\right)}{N\,\hat{p}\left(w_{n},y_{n},z_{n}\right)}=\frac{\hat{p}\left(z_{n}|w_{n},y\right)}{N\,\hat{p}\left(z_{n}|w_{n},y_{n}\right)}
\end{eqnarray}

\section*{Appendix B: Details of experiments\label{sec:Appendix-B}}

\addcontentsline{toc}{section}{Appendix B: Details of experiments}Each
sample is indexed by the \emph{environment }or sample variable $e\in\left\{ 1,2,3\right\} $.
All prediction estimates are replicated over 10 runs, obtained by
either synthesising entirely new data, or randomly permuting the observations
before splitting into distinct samples. MATLAB code which implements
these experiments is available on request from the authors.

\subsection*{Synthetic Gaussian mixtures}

The target and mediator variables take on values $\Omega_{Y},\Omega_{Z},\Omega_{U}=\left\{ 1,2\right\} $.
The model is:

\begin{eqnarray}
U & \sim & Bernoulli\left(p\right)\nonumber \\
Y|U & \sim & Bernoulli\left(q_{e}\left(u\right)\right)\nonumber \\
Z|Y & \sim & Bernoulli\left(r\left(y\right)\right)\\
X_{1}|Z,U & \sim & \mathcal{N}\left(\mu_{1}\left(z\right),\sigma=1\right)\nonumber \\
X_{2}|Z,U & \sim & \mathcal{N}\left(\mu_{2}\left(u\right),\sigma=1\right)\nonumber 
\end{eqnarray}
The parameters depend upon the sample and whether the model is back-door
or front-door. For back-door, we set $p=0.85$. The parameter $q_{e}\left(1\right)=0.95$
and $q_{e}\left(2\right)=0.05$ for $e=1,2$, and we set $q_{3}\left(u\right)=0.5$
(thus making the target independent of the confounder for sample 3).
The mediator plays no role in this model and so $r\left(1\right)=1$
and $r\left(2\right)=0$. The feature parameters are $\mu_{1}\left(1\right)=1.5$
and $\mu_{1}\left(2\right)=-1.5$, and $\mu_{2}\left(1\right)=2.4$
and $\mu_{2}\left(2\right)=-2.4$.

For front-door confounding, we have $p=0.5$, for $e=1,2$ we set
$q_{e}\left(1\right)=0.98$ and $q_{e}\left(2\right)=0.02$, and $q_{3}\left(u\right)=0.5$
(non-confounded sample). The mediator has conditional parameter $r\left(1\right)=0.90$
and $r\left(2\right)=0.10$. The feature parameters are the same as
for the back-door case.

\subsection*{Background-MNIST data}

In this example, we embed and modify the 28 by 28 pixel digit image
data from the MNIST dataset \citep{lecun_mnist_2019} inside a back-door
or front-door causal graph. We use the 2,000 test images for digits
`2' and `6'. For convenience, the back-door samples are generated
using the following model (note that we can always find the equivalent
conditional $Y|U$ using Bayes'):
\begin{eqnarray}
U|Y & \sim & Bernoulli\left(q_{e}\left(y\right)\right)\nonumber \\
Y & \sim & Bernoulli\left(p\right)\\
\boldsymbol{X}|Y,U & \sim & MNIST\left(y,u\right)\nonumber 
\end{eqnarray}
where $p=0.5$, and $q_{e}\left(1\right)=0.95$, $q_{e}\left(2\right)=0.05$
and $q_{3}\left(y\right)=0.5$ (non-confounded sample 3). Here, $MNIST\left(y,u\right)$
is a random function which retrieves a unique MNIST image $\boldsymbol{x}$
representing digit `2' for $y=1$ and digit `6' for $y=2$. The the
brightness of the image data is then modified by converting $\boldsymbol{x}\mapsto\min\left(\boldsymbol{x}+\boldsymbol{b}\left(u\right),255\right)$
where $\boldsymbol{b}\left(u\right)$ is a 28 by 28 image of pixels
all with the same value 100 if $u=1$, and value 0 otherwise.

For the front-door case, the model is:

\begin{eqnarray}
U & \sim & \mathcal{N}\left(0,5\right)\nonumber \\
Y|U & \sim & Bernoulli\left(q_{e}\left(u\right)\right)\\
Z|Y & \sim & Bernoulli\left(r\left(y\right)\right)\nonumber \\
\boldsymbol{X}|Z,U & \sim & MNIST\left(z,u\right)\nonumber 
\end{eqnarray}

Here, the Bernoulli parameter for $Y$ depends upon the confounder
in the following way: 
\begin{equation}
q_{e}\left(u\right)=\frac{q^{u}}{q^{u}+\left(1-q\right)^{u}}
\end{equation}
where $q=0.8$ for $e=1,2$ and $q=0.5$ for $e=3$ (making the target
independent of the confounder for sample 3). The mediator parameter
is $r\left(1\right)=0.95$ and $r\left(2\right)=0.05$. As above,
$MNIST\left(z,u\right)$ is a random function which retrieves a unique
MNIST image $\boldsymbol{x}$ representing digit `2' for $z=1$ and
digit `6' for $z=2$. Then, the brightness of the image data is modified
by converting $\boldsymbol{x}\mapsto\min\left(\boldsymbol{x}+\boldsymbol{b}\left(u\right),255\right)$
where $\boldsymbol{b}\left(u\right)$ is a 28 by 28 image of pixels
all with values:

\begin{equation}
v\left(u\right)=100\times\left(\frac{1}{2}\arctan\left(\frac{1}{5}u\right)+\frac{1}{2}\right)
\end{equation}
which maps the continuous confounder values onto the scaled background
brightness values in the range $\left[0,100\right]$.

\subsection*{Parkinson's voice data}

This dataset is based on features extracted from the sustained phonations
of three sets of patients with Parkinson's disease and healthy controls,
recorded in separate labs based in the US \citep{little_parkinsons_2008},
Turkey \citep{sakar_parkinsons_2018} and Spain \citep{naranjo_parkinson_2019}.
For further details of how these samples were recorded, see references
therein. Features which are common across all three datasets are extracted,
to give a feature dataset $\boldsymbol{X}$ of $9\times1191$, along
with a target class from $\Omega_{Y}=\left\{ 1,2\right\} $ for each,
representing healthy control versus Parkinson's. The goal is to predict
whether any single observation is from an individual with or without
Parkinson's.

The confounder $U$ takes values in $\Omega_{U}=\left\{ 1,2,3\right\} $
denoting the dataset. For illustrative purposes, for the confounded
environment samples 1 \& 2, the effect of different labs, patient
populations and experimental protocols is enhanced by resampling with
replacement from the combined dataset to ensure that the confounding
across datasets, conditioned on the target variable, is in the proportion
5\%, 5\%, 90\% for $Y=1$ and 90\%, 5\%, 5\% for $Y=2$, whilst simultaneously
ensuring that the target variable is balanced across classes. For
sample 3, the proportion of data from each dataset is 47.5\%, 5\%,
47.5\% independent of $Y$ (so that sample 3 is non-confounded).
\end{document}